# Adding Context to Concept Trees

Kieran Greer, Distributed Computing Systems, Belfast, UK.
http://distributedcomputingsystems.co.uk

Version 1.3

***Abstract –*** A Concept Tree is a structure for storing knowledge where the trees are stored in a database called a Concept Base. It sits between the highly distributed neural architectures and the distributed information systems, with the intention of bringing brain-like and computer systems closer together. Concept Trees can grow from the semi-structured sources when consistent sequences of concepts are presented. Each tree ideally represents a single cohesive concept and the trees can link with each other for navigation and semantic purposes. The trees are therefore also a type of semantic network and would benefit from having a consistent level of context for each node. A consistent build process is managed through a 'counting rule' and some other rules that can normalise the database structure. This restricted structure can then be complimented and enriched by the more dynamic context. It is also suggested to use the linking structure of the licas system [15] as a basis for the context links, where the mathematical model is extended further to define this. A number of tests have demonstrated the soundness of the architecture. Building the trees from text documents shows that the tree structure could be inherent in natural language. Then, two types of query language are described. Both of these can perform consistent query processes to return knowledge to the user and even enhance the query with new knowledge. This is supported even further with direct comparisons to a cognitive model, also being developed by the author.

## 1 Introduction

A Concept Tree is a structure for storing knowledge where the trees are stored in a database called a Concept Base. It sits between the highly distributed neural architectures and the distributed information systems, with the intention of bringing brain-like and computer systems closer together. Concept Trees can grow from the semi-structured sources when consistent sequences of concepts are presented. Each tree ideally represents a single cohesive concept and the trees can link with each other for navigation and semantic purposes. The trees are therefore also a type of semantic network and would benefit from having a consistent level of context for each node. Concept Trees were introduced in [9] for the purpose of creating structure over possibly heterogeneous and unrelated data events. With a model that is closer to the human brain, time-based events were considered, where time would be the sequence indicator and each event would be a single concept to be stored. This sort of information cannot be easily modelled beforehand and so the Concept Base grows dynamically from it, using some very basic and generic rules to do so. Testing so far however has been on text documents, where the structure inherent in text is used to build the concept trees. That paper noted difficulties in adding context automatically, which is a problem that this paper attempts to address.

The term 'concept base' has been used previously ([13] or [23], for example) and has been adopted in [10] to describe a database of heterogeneous sources of arbitrary concepts. They would be more similar to an Object-Oriented or Graph database than a Relational one [3][19], for example. The term 'concept' can be used to describe a single value or a complex entity equally and so the concept base can store information of arbitrary complexity. If the information is arbitrary, it is probably the case that some level of structure must be added first, before the information can be processed, data mined, or reasoned over. One realisation of the concept base is as a flat structure of 1 or 2 levels [10] and another is as concept trees. The trees start with a root node that is extended by branches, where a strict 'counting rule' does not allow a branch node to have a higher occurrence count than the parent node. A formal and mathematical process for doing this has been described [9], allowing potentially unrelated concept groups to build the tree-like structures. The trees can be used to store image or behaviour objects [6] equally, when the latter might be compared with Markov Models [5]. To give a difference then, possibly state changes should occur across trees and not inside of any tree, when the tree remains as a single cohesive concept. Therefore, if a tree produces a backlink to another part in itself, that part should be split-off to create a new tree. The backlink would then become a link to the new tree and traversing it would cause a state change. An exception to this may be a backlink to the base of the tree, which simply means that the tree should repeat.

Even if the concept that a tree represents is quite abstract, a consistent build process would mean that it can be used and shared in a consistent way. The tree structure can therefore be the basic building block for a distributed database and it can also be normalised. A test of parsing scientific papers and building trees from them shows that the counting rule is probably inherent in natural language. This paper extends the theory by adding layers of descriptive elements to each of the tree nodes. The nodes form a more permanent structure that may change slowly over time, but descriptors can add context, when a more dynamic type of process across the structure can occur. With considerations for context, it is possible to extend the mathematical model further, thereby allowing the construction process to be automated in that respect. Appendix A gives a full list of the rules and equations that have been suggested for building a concept base, including the new work of this paper.

The rest of this paper is organised as follows: section 2 gives some related work. Section 3 reviews concept trees and introduces some new ideas, along with section 4 that extends the mathematical model for context. Section 5 runs some tests on global tree properties, while sections 6 and 7 describe some query language options for the new model. Finally, section 8 gives some conclusions on the work.

## 2 Related Work

The Concept Base is a type of distributed database and can be compared to the other new database types. This has been described already in [9], where that summary is repeated again next: The recent problems that 'Big Data' provides, linking up mobile or Internet of Things with the Web, has meant that new database structures, or particularly, their indexing systems, have had to be invented. Slightly more akin to Object-oriented databases are new database versions such as NoSql and NewSql [12], or navigational databases. As stated in [12], the modern Web, with the introduction of mobile and sensor devices has led to the proliferation of huge amounts of data that can be stored and processed. While the relational model is very good for structured information on a smaller scale, it cannot cope with larger amounts of heterogeneous data, as it is usually required to process full tables to answer a query. NoSql now has different meanings and might also be termed 'Not Only SQL'. It can use different indexing systems that might not even have an underlying schema. So it can be used to store different types of data structure, probably more as objects than tables. The database aspect however can try to provide an efficient indexing system, to allow for consistent search and retrieval over the distributed contents.

Graph databases [19] originated from graph theory and use graphs as their data model. By using a completely different data model to the other 3 types, graph databases can efficiently store the 'relationships' between different data nodes. Graph databases are specialized in handling highly interconnected data and therefore are very efficient in traversing relationships between different entities. A navigational database is a type of database in which its records or objects are found primarily by following references from other objects. Navigational interfaces are usually procedural, though some modern systems like XPath [22], can be considered to be simultaneously navigational and declarative. Navigational databases therefore use a tree indexing system and can fall under the graph-based category of NoSql. While the problems of semi-structured or unstructured data remain, these new databases do offer general architectures and indexing systems. One criticism of graph-based ones however, is that they tend to lead to very messy sets of indexing links that do not have very much structure. This is possibly the case for concept trees as well, but as the concept tree might have a more ordered construction process, it can provide some kind of mathematical foundation to help with the organisation.

This paper adds context to the tree structure, which was missing from the first version, written about in [9]. However, the need for context was clear and again as quoted from the first paper: For dynamic or autonomic systems, the context in which the knowledge is used can become a critical factor. Context is an information space that can be modelled as a directed graph, rather like an ontology. Context allows both recognition and mapping of knowledge, by providing a structured and unified view of the world in which it operates [1]. It is about evolving, structured and shared spaces that can change from one process to the next, or even through the duration of a single process. As such, the meaning of the knowledge will evolve over time. Contexts are defined by a specific set of situations, roles, relations and entities. A shift in context corresponds to a change in the set of entities, a change in the set of possible relations between entities, or a change in the set of roles that entities may play.

The Concept Tree has been compared to Markov Models [18][5] and is even similar with respect to the counting rule that can be implicit in a Markov Model but does not have to be. The paper [21] is a survey of these and notes that there are many different types that obey different rules. As explained in the Introduction, a concept tree can be made different by not allowing backlinks between nodes internally, but only across trees. In that case, a tree would be compared to a single Markov Model node. If parsing text to produce word sequences, then NGram structures [2] are also related, but they are typically used to predict word sequences, which is a more exact measurement than a type of set membership. Another new structure is a Graph Convolutional Network [14], which is an example of layers of features, but that is maybe going too far in the neural network direction. As the concept tree may be used as an indexing system, every node in it is understood and so it would not have the black-box properties of a neural network either.

This paper is based on the earlier papers by the author [8]-[11]. In particular, [9] introduced the idea of concept trees, while [10] describes a related symbolic neural network and the query language that is to be used. The book [11] writes about the original linking ideas, including the permanent and dynamic links that are now part of the structure. The software has been written with the help of the licas system [15] and also OpenNLP [17] and WordNet [4][16]. Licas is an open source framework written in Java, for building distributed service-based networks. It includes a number of AI algorithms, including some new ones. It also includes a number of tree nodes and structures that the concept trees can be built from. A Concept Base [10] is the idea used for the database that manages the concept structures. In one type of implementation it can be relatively shallow or flat and in another it can manage the deeper concept tree structures of this paper. Two earlier examples of other concept databases include [13] and [23]. As the concept base is also a type of semantic network, one reference for that would be [20].

## 3 Concept Trees

Concept trees [9] start with a base node that is extended by branches of related nodes. A strict counting rule does not allow a branch node to have a higher number of occurrences than the parent node. This is actually the case if the tree is always updated starting from the base, but might not be the case if the tree can be updated from other places. As part of the process therefore, trees can be split into two or more, where links between the parts can be maintained, to help with navigation across the trees. The paper [9] and Appendix A give a more formal set of rules that would build the database in a normalised way, and also how to update and re-structure it. Figure 1 is an example of what a concept base might look like with 3 trees.

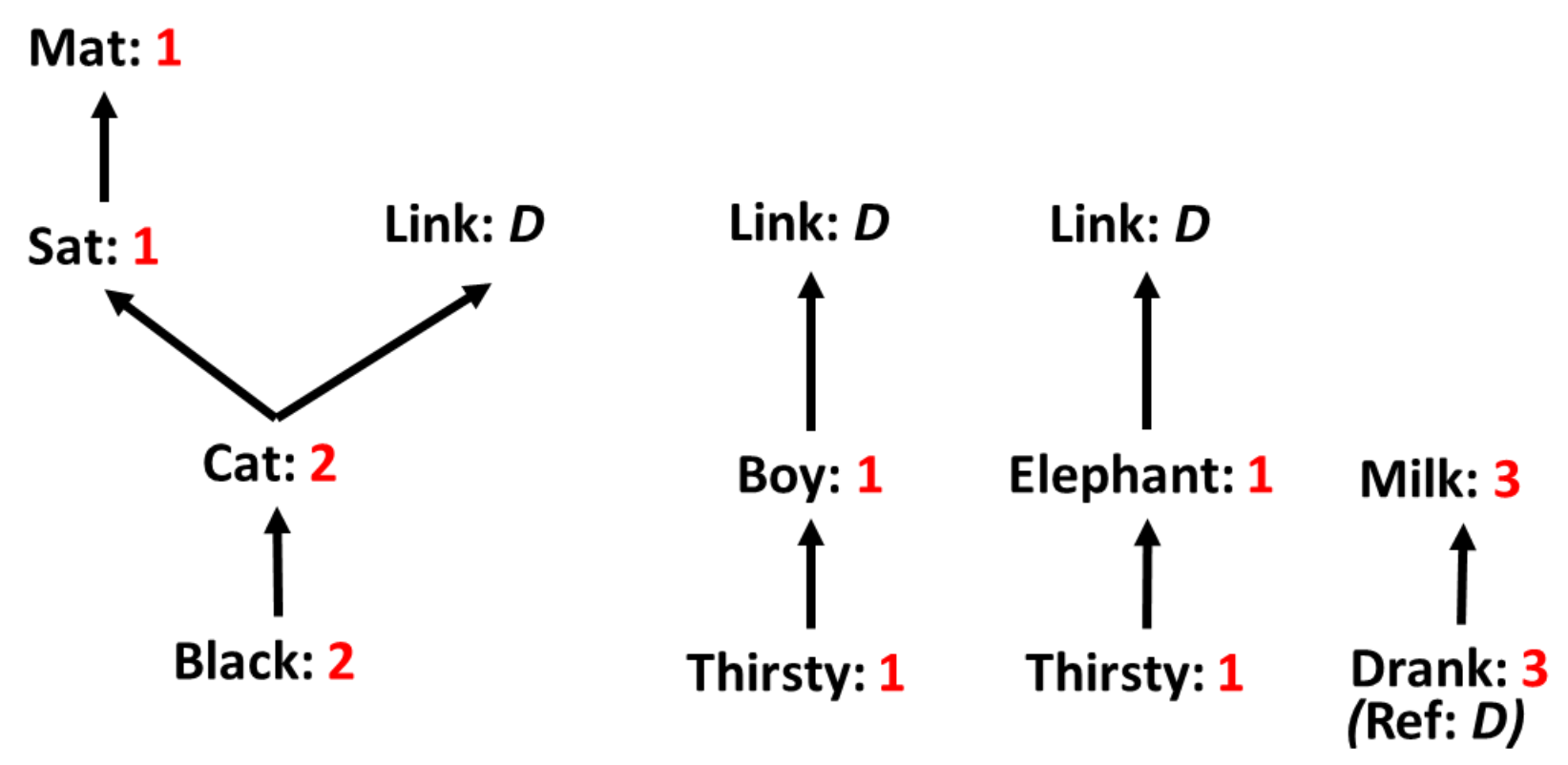

Figure 1. Example trees with indexed links for navigation.

Because ‘drank milk’ is important to more than one tree, it creates a separate tree that gets linked to (key *D*) by the other concepts, for example. Each tree is also a distinct concept. In this case, drank and milk are considered to be a single concept. If another food is drunk, then that tree structure would change. Note also that the first 3 trees have descriptive base elements. This seems a bit unnatural and ‘cat’, ‘boy’ and ‘elephant’ should probably be the base elements. It is very unlikely that the current set of base elements can maintain themselves and the tree structures would surely be changed when new information is added. These descriptive elements should therefore be moved to a more dynamic context, as described in section 3.2. An earlier version of this paper considered using the tree as a process, with backlinks inside of a tree to be used as process loops. As written in section 1, this possibly over-complicates the problem and is better handled by the behaviour scripts written about in [6].

### 3.1 Creating Trees from Text Streams

One option would be to create trees directly from text streams, where parsed word sequences can build the tree structure. There are different ways to parse the text stream, including a statistical evaluation first of the most popular words, to be used as base nodes; or possibly simply parsing each sentence separately and adding as is. Natural language is already highly structured and so using as is would pass some of that structure onto the concept base. To demonstrate this, section 5.1 gives the results of some basic tests that parse text documents into concept trees.

### 3.2 Contextual Descriptors

Figure 1 showed that the descriptive elements do not sit naturally as part of the static tree structure, but adding context would allow for a richer type of information. In the first instance, context can be used to define appropriate paths through the trees. The context would still be dynamic and could include a weighted calculation, where a correct path one day may not be correct the next day. So this can still result in tree parts to reason over. A new suggestion for this paper therefore, is that the tree nodes should only be nouns or verbs. These are more solid and static real-world concepts, as opposed to the more transitory descriptive ones. The nouns and verbs would therefore make up the static tree structure. Each node would then be allowed to have any number of descriptors – adjectives for nouns and adverbs for verbs. For simplicity and a clear process, the descriptors link only with other descriptors.

Figure 2 is a version that has added a layer of descriptors, one for each node. The black lines are the tree structure, while the blue lines are from the concept to its descriptors and the dashed lines link descriptors with each other, probably created through dynamic feedback.

As shown in Figure 1, there is a black cat, but the cat could also be white, tabby, or whatever. Also, water might be available to drink as well as milk. So the context in Figure 1 should more correctly be changed into the descriptive elements of Figure 2 and added as dynamic content to the related node. If the context nodes are more transitory, then they do not have to follow the strict counting rule of the static structure and so they can build more arbitrary relations with other descriptors in the same tree or even across trees. This more dynamic structure would also relate to cognitive processes.

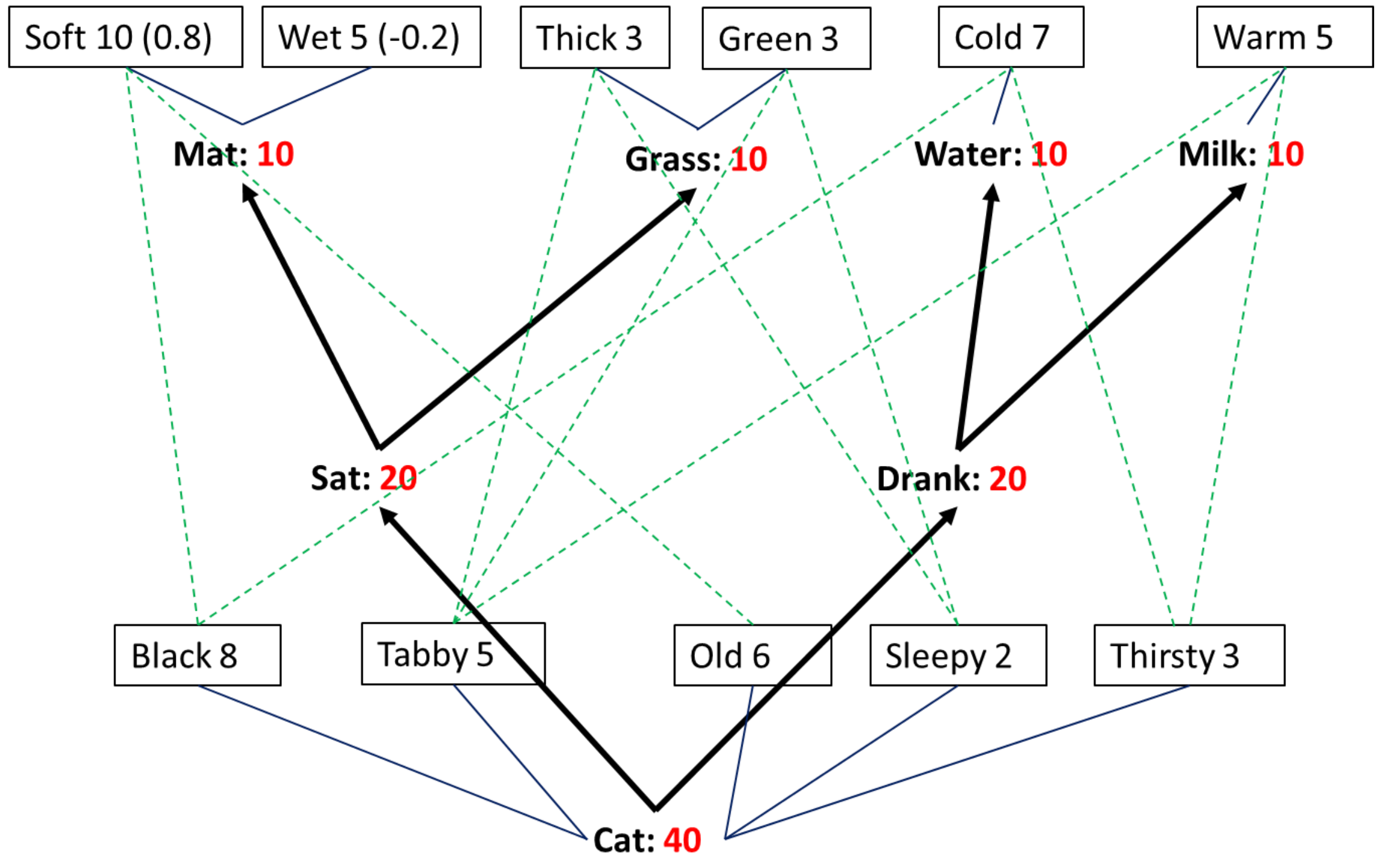

Figure 2. The Concept Base Trees with Descriptors

### 3.3 Weight Reinforcement

Just to show the possibilities, the descriptors have occurrence counts, but they might also have weight values for positive or negative reinforcement. For example, a soft mat would reinforce the path value with a weight value of 0.8 (10 x 0.8), but a wet mat would reinforce it negatively with a weight value of -0.2 (10 x -0.2) and therefore suggest a different path through the tree if that realises a larger total count. There are therefore possibilities for positive and negative reinforcement that could occur automatically when the system is updated. Section 4.5 gives some rules about when weight values should be updated.

# 4 Extending the Mathematical Model

This section extends the mathematical model of [9] to take consideration of a more dynamic structure. Some ideas have been based on nature and with the introduction of equations, as well as the counting rule, dynamic behaviour can be realised.

## 4.1 Word Overlap

This is the case where single words or concepts exist in more than 1 tree. This can happen as part of the construction process and it may be easier to allow a particular concept to exist more than once, rather than enforce that the node should be removed to create a new tree. Especially if the node is simply part of a path and not particularly important.

## 4.2 Tree Shape

Tree size, shape and the idea of the concept ordering, can also be considered mathematically and is helpful when considering how to link trees together or join two trees. One factor is the fact that a larger tree, representing a more stable concept, should not join as a branch to a smaller tree, but the counting rule would probably prevent this. The shape of the tree is also significant. A wide shallow tree is more likely to be a base tree than a narrow deep tree and it should be easier to add to the wide shallow tree. What a node represents can even be considered ([9], section 6.4), such as a car drives on a road and so always have the road tree at the bottom, even if it is a smaller tree. This could be indicated by lots of links to other trees from it, or additional knowledge, from an ontology, for example. A function of factors that therefore include the size of the tree (number of nodes $n$) and its width and depth ($w$, $d$) can help to define the tree itself. Although, if the width and depth are related, then only one of them is required with the number of nodes, as in:

$fs_t$(n, d). Equation 1

### 4.3 Joining Trees

The first paper [9] considers that a tree-join or re-join is a more intelligent act than breaking a tree up. Intelligence introduces some level of control and while a narrow deep tree has more entropy than a wide shallow one, the additional intelligence, in the form of an equation, would override the more automatic process of the narrow tree breaking into a set of shallow ones. The trees would have larger frequency counts inside of each than in the links between them and are also separated by the counting rule. As this is mandatory, the difference in the frequency count size could help to indicate if trees should be re-joined as a single tree concept. If a link's count, for example, becomes closer and closer to the base count for the tree it is linking with, then this might suggest to join the two trees. One definite rule can be:

Lemma 11:

*The tree join is legal, if the link is the only link to the base of the second tree. In which case normalisation is maintained. If that is not the case, then an equation can be used.*

If this is not the case, then it is also interesting to consider the tree shapes. Another consideration is if the tree nodes share similar context. If the context is very different then maybe they are separate instances of similar objects and it would not be a good idea to put all of the context together. So, it would be possible to model this with a mathematical equation that indicates when it is legal and when it is not legal. Therefore, if these are the main parameters, then if comparing two trees and considering if they should be joined: the two shape functions can be considered along with the branch link (count) in tree 1 ($l_{t1}$) and the base node link (count) in tree 2 ($l_{t2}$) and finally any concepts (context) in either tree ($C_{t1}$ and $C_{t2}$), as in:

$f_{join}(fs_{t1}, fs_{t2}, l_{t1}, l_{t2}, C_{t1}, C_{t2})$. Equation 2

### 4.4 Where to Add Context

Adding context can also be managed through some basic mathematics. For example, if we have a tree with the concept ‘road’ at the base, then there may be lots of shallow branches for vehicles like ‘car’, ‘lorry’, ‘bus’ and so on. The tree can define that the vehicles drive on the road, but it is not very interesting to know that a blue, red, or white car drove on the road, for example. This context has no meaning and belongs somewhere else. Therefore, the context that gets added should relate directly to a link in the tree, although this can be simply the link from the parent node. Another way to look at this is that the context will become an active factor in the tree search. It will have counts and weights that are updated when the tree is traversed. Possibly a decay factor would remove it over time, but it may be possible to be more specific and state that:

Lemma 12:
*A context should only be added if it would potentially change the search path in a tree*.

Linking between trees however may also want to consider the context and require that the branch and base nodes share some context to allow the link, or at least, do not clash on context. For this example, it is therefore likely that each individual car tree would have the richest context additions and that the counting rule would also require them to create their own separate trees. The road tree would link to them based on structure only and would not store all of the different contexts itself, unless the road tree was itself a specific instance with a particular ‘road’ context. Therefore, it is valid to store context with each concept and add this to the concept base in concept-context pairs [10], as in:

[A1, (Va1,Va2)] [B1, Vb1] [C1, (Vc1,Vc2,Vc3)] Equation 3

In this instance, A, B or C is the concept type and Va1, Vb1, Vc1 is a related context value. If the concept base stores information even as separate pairs, then the context can help to define where a concept should go. If the source can also provide information on links, then that will help, as in the text parsing of section 5.1.

### 4.5 Count Updates

If the tree nodes are always updated from the base up, then the counts will remain consistent. If context is used, then this is not static structure and so it may be preferable to update it separate of the tree structure. Therefore, the context gets added and updated only during use, which can be positive or negative feedback. The tree structure can be changed when new nodes are added or again through use, but that is the final path selection and not the search process. Two rules might be:

Lemma 13:
*When searching over trees, if a context is true, it can be updated, but node counts should not be. When selecting a tree path or changing the tree structure, tree nodes can be updated, but context nodes should not be.*

Lemma 14:
*If a context is true but the node with it is not selected, then a weight update should be negative. If the context is true and the node with it is selected, then a weight update should be positive.*

## 5 Concept Tree Tests

Some tests have been run on global tree properties and also to determine how close a match they are to regular text. Later sections then consider some possibilities for using the trees as part of a query process.

### 5.1 Parsing Text Documents

When writing or speaking, it is typical to introduce an item before describing it. The description comes after the introduction and if the description also repeats, then it can have some statistical meaning. One option therefore is to create new trees from text streams, where parsed word sequences can build the tree structure. For this set of tests, popular words were determined by parsing the document and counting the frequency of each word, where the 10 most frequent words were used as base words for nested sequences. This would be compatible with the counting rule of concept trees and with a continuous text stream, maybe a start or end is not very clear. Common words, or words of length 2 or less were also removed. In fact, the most popular word overall would be the base node in the tree and the other 9 words would become child nodes at level 1.

Each base keyword would then realise a sequence from one occurrence to the next and in that sequence, the 10 most popular words would be kept as the nested sequence words. Note that the ‘10’ value was arbitrary and could be anything. This would produce a number of related word paths over the whole document for each base keyword. The tree was then constructed by adding the keyword at the base and the nested set as child nodes, where branches would be created when sequences overlapped. The nested nodes would actually be added at the first possible position in the tree, not necessarily a child of the current position, if the node already existed, again to be more compatible with the counting rule.

## 5.2 Text Parsing Tests

While popular word counts is the main factor, because sequences are also determined by relative positions, a better distribution in the sequences can result in a slightly less frequent word obtaining a larger count when the sequences are aggregated together. When that happens, the tree may need to be split because of the counting rule. This happened only three times in 75 documents however. The trees most often had a depth of 2 (root concept and nested concept), and 3 less often, but deeper trees with a depth of 4 or more was possible. This is in fact a curious aspect of putting the most popular words at the base. Simply parsing text sentences, which was done for the tests in sections 6 and 7, produced much deeper trees, but also duplicated word terms sometimes. It seemed easier for those tests to keep some duplication and maybe suggests that using popular words gives the best normalisation. With regard to document overlap, there were 234 nodes anywhere in more than 1 tree compared to 686 nodes anywhere in just 1 tree. Of those 234 nodes, 52 were at the base of a tree and 39 of those trees would be linked to from other tree branches, which is over 70%. There were 82 separate trees in total, created from the 75 documents, an increase of less than 10%. This construction process therefore appears to show that the counting rule is inherent in the text structure and so the reduced tree content could give some type of summarised meaning of the related text document.

### 5.3 Tree Shape Tests

The book [11] was divided into separate chapters and the variance and tree shape measured for each chapter. Variance is a measure of difference in the numbers of each word, which might be relevant to a structure that uses a counting rule. While the word variance could vary quite a bit, the tree shape stayed almost the same. The shape calculation was:

$$\text{Shape} = d / n. \qquad \text{Equation 5}$$

Each chapter produced several trees and so this was averaged for all trees with 10 or more nodes. After common words were removed, the word variance ranged from 2.5 to 115, but the tree shape ranged only from 0.48 to 0.66. However, the two smallest variances gave the two largest shape values and the two largest variances gave the two smallest shape values. This might again suggest that the text structure is relevant and even if word counts vary widely, the natural language structure will still provide some order that the concept trees can make use of.

## 6 First Query Language

As the concept tree is a type of database, it should be possible to use it to query for information. A test program was written in the Java programming language. It does not implement the concept base in full, with regard to all of the rules, but it allows for trees to be added and then searched over. The program can read a text document, parse it into sentences, remove the additional formatting and then use OpenNLP [17] to construct each phrase or word group. WordNet [4] can then be used to determine the concepts with related descriptors. A most basic query can take the form of a tokenised list of words that can match with the general structure. With tokenised word sequences, a Horn clause [10] might be appropriate and would keep things relatively simple, although there are several configuration parameters that can change the results. In this example, each clause in the query was a tree concept with a related descriptor and the query process would try to find all trees that match with the terms. If a slot in the query clause was empty, then a returned tree could make a suggestion for the missing information.

### 6.1 Tree and Query Construction

As an example, an NLP query was parsed into associated word phrases, where each word phrase would be presented as a single statement to the concept base. Nouns and verbs need to be recognised first. Then, if a noun has an adjective either side of it, that can be added as a descriptor for the noun. If a verb has an adverb either side of it, that can be added as a descriptor for the verb. So the NLP might add some ambiguity, but it would still have to match with the database and that would help to confirm what is correct. For example, the descriptor could be added anywhere it might be relevant and would only be used if it was in fact returned as part of a search process.

### 6.2 Test Scenario

For the query, text sentences were parsed in the same way as for constructing the trees and used to search over the concept base. The query process is required to return some knowledge or information and so when a query construct was executed over the linked database, instead of updating links, it could fill-in missing slots in the query clauses. If different tree sets return a choice of answer, then some reasoning might be required, but there are a lot of numerical values to suggest what the best answers might be and so there would be a lot of help for evaluating the best reply. As the database links also get updated, ambiguity in the query can eventually be removed, if it does not result in successful matches with the tree structure.

### 6.3 Test Example

A very simple example would be parsing the sentence ‘Jack wore a white shirt and blue trousers’. If this was added to the concept base and a query executed asking about [shirt-white] $\wedge$ [trousers-?], then the tree structure would return [shirt-white] $\wedge$ [trousers-blue], suggesting that the trousers are blue. Equally however, a question like ‘was something blue worn with a white shirt?’, or [shirt-white] $\wedge$ [?-blue], could be asked and answered.

# 7 Second Query Language

A second query language can make use of the concept tree structure to return results that may also be based on sentiment. This is only one example of many possibilities of this type, but it is another example of adding new knowledge to the query process. The tree nodes represent word terms, where each node also has a set of descriptor terms. Those descriptor words however are also added as regular tree nodes. The search process could then use the descriptor elements to link with the tree nodes with the same value. This would mean that a query reply could return information about objects not directly asked for, but that were related through the descriptor elements, or some type of sentiment. Another option would be to link two nodes through having similar descriptor elements.

## 7.1 Test Scenario

In this example, the query took the form of a list of words and a path length to find those words in. The trees would be searched for matches and would allow any path words in-between to be included, as long as a maximum path length was not exceeded. Using these paths, additional words at the end could be included if again, they did not exceed the maximum path length. One query term would therefore give the most general type of query and return the most potential paths. This would produce a list of word sequences to find in the source text, in a certain order, but not necessarily one directly after the other. Then to add new knowledge to the query, descriptor elements for the query terms could be retrieved and matched to tree nodes as well. If a match was found, the query term could be replaced by the descriptor and the trees re-searched, to produce another set of query paths. The full set of query terms would then be matched to the source text (that the trees were created from) and matching sentence parts returned as the result.

### 7.2 Test Example

This example is for the third chapter of the book [11] that is about knowledge and networks. The query term used was 'network' and this had a descriptor element about 'knowledge-based', which did map to a knowledge-based tree node. Therefore, as well as returning source text that was about networks, the query would also return source text that was about knowledge-based, decided by the system itself. The text sequences shown in Figure 3 were therefore returned for the 'network' query.

1. independently conceptualised and developed network systems that used datagrams, or packets of information, that could be used in a packet switched network between computer systems.
2. created the first wide area network (WAN)
3. These have similar aims to neural networks, where a distributed network of simpler components tries to achieve some greater goal.
4. notes his own efforts in this area with his PhD work on building a computer chess program that uses a neural network as part of its search heuristic
5. Berners-Lee recognised that for the Internet to evolve, it would need to be turned into a knowledge-based network.
6. Researchers also looked at the neural solutions, which may be able to overcome the obvious problems of the knowledge-based approaches.
7. This could also be considered as combining the knowledge-based and the experience-based approaches.
8. function, it may have its own internal knowledge-based algorithm that can perform autonomous or independent behaviour and make intelligent decisions
9. If considering the knowledge-based/experience-based divide, then symbols and logic fall into the knowledge-based category, while neural-based and statistics fall into the experience-based one.

Figure 3. Search Results returned for the 'Network' query with 'Knowledge-based' descriptor.

The 'Berners-Lee' sentence introduces the relation between networks and knowledge, which could make the reader interested, and then some later sentences only mention the knowledge aspect. The exact structure of the returned sentences is not important here, because this is only a prototype system.

### 7.3 Search Oscillation

Sentences in the source text that contained any of the query sequences would be returned as the result, where another option would be to take those sentences only and re-create concept trees from them. If the query search was then run over the new tree structures, it did not return exactly the same result. This is because the new trees were not created from exactly the same source text. However, if this process was repeated 3 or 4 times, then eventually the adjustment stabilised to a set of trees and word sequences that did not change. It would be dangerous to compare this too closely with the ensemble-hierarchy structure of [6] or [7], but it might be interesting because it uses a similar type of oscillating cycle between the two structures. As the test prototype re-creates the trees from the source text each iteration, this would have to be made more economic for a distributed search engine, but the potential for a new query process is clear. One option would be to allow a base node to link to any of its nested nodes, so that it could skip levels if required.

## 8 Conclusions

This paper describes enhancing concept trees with layers of descriptive context. This provides a richer level of knowledge and would make them more useful in practice. The context can be used, both to direct the search process and also to suggest tree structure (when nodes across trees should be linked). It can probably be used for much more than this, but a for a first model, selecting the correct paths through the trees should be enough. It has also been possible to update the underlying mathematical model and suggest some new rules for the context. The tree structure is now both static and dynamic, with different elements being updated in different phases of the search process. The descriptor elements can also be weighted and have a positive or a negative effect. Basic tests on parsing text documents show that they probably inherently contain the concept tree structure. While the structure, for example, may be contained in any type of nesting, natural language uses it implicitly and the concept trees might even be a summary of the parsed text document.

For the database model, the query structure of a Horn clause [10] can be tried, where the idea of an attribute name and value is the same as the concept name and context of this paper. In [11] query feedback over a distributed network was used to update links and a similar scenario is realised in this paper. A second type of query process is also interesting, where it can add new knowledge through the descriptive elements, possibly related to sentiment. This could result in replies making statements that the original query would not have thought about. It also demonstrates an oscillating and stabilising cycle between the source text and the concept trees that can be compared to the ensemble-hierarchy structure, written about in [6] and [7]. Because that work is still quite abstract, a close comparison is probably not a good idea, but if the descriptive element is the glue between two concepts, for example, then this has a striking similarity to how the behaviour script in [6] is built. It could suggest that the third element in each behaviour clause of that paper is in fact the context or situation that binds the clause together.

## Appendix A – List of Rules and Equations for the Concept Base

This appendix gives a full list of rules and equations for the concept base construction and use, taken from this paper and [9].

### A1. Rules

There are some additional points in [9], section 5, but this is a full list of the main rules.

Lemma 1

A concept tree can represent different types of entity. The entity however should be a whole concept. For example, it might be a single object in the real world, or a single process.

Lemma 2

The base concepts in any tree are the ones that should be used first in any scenario.

Lemma 3

An earlier paper gave a slightly philosophical argument that if two concepts are always used together, then at some level they are a single entity. If concepts are always used together, then they can always be part of the same tree.

Lemma 4

The 'counting rule' states that the tree structure requires that every child node has a count that is the same or less than its parent.

Lemma 5

For a tree to exist, every node in it must be true. That does not mean that every node is evenly used, but there should be no false information. This extends to being true for any entities that link to the tree or related sub-tree. If any part is false for any linking entity, then the tree needs to be split. This can be a positive or negative truth however and can help when deciding about context.

Lemma 6

Adding a new group of concepts should always start from the base node: (a) If it would start part of the way up a tree, then the process should form a new tree instead. (b) Similarly, when a new path is added to an existing tree, it should start from the base only and traverse through any sub-path up to any leaf node. It can then extend that leaf node if required.

Lemma 7

Tree comparisons and updates are made using groups of concepts that represent individual input events. The event is considered to be a complete entity itself and is compared with the other structures as that, where it needs to match with existing tree paths in one of two ways: (a) If it matches exactly from the base of another tree up any branch, then it can be added to that tree. (b) If its' base matches to a different node of another tree, then a link between the two trees can be created.

Lemma 8

Linking related nodes is always possible. Different keysets can then define, for different entities, what links are traversed.

Lemma 9

To differentiate from Markov Models, state changes should occur across trees and not inside of any tree. Therefore, if a tree produces a backlink to another part in itself, that part should be split-off to create a new tree. The backlink would then become a link to the new tree and traversing it would cause a state change. An exception to this may be a backlink to the base of the tree.

Lemma 10

Putting a heavier object on-top of a lighter one is not often done and so there is a natural order here. This ordering, based on some external knowledge of the real world, might also become part of the structure.

Lemma 11:

The tree join is legal, if the link is the only link to the base of the second tree. In which case normalisation is maintained. If that is not the case, then an equation can be used. Re-joining trees needs to consider the base entity links more.

Lemma 12:

A context should only be added if it would potentially change the search path in a tree.

Lemma 13:

When searching over trees, if a context is true, it can be updated, but node counts should not be. When selecting a tree path or changing the tree structure, tree nodes can be updated, but context nodes should not be.

Lemma 14:

If a context is true but the node with it is not selected, then a weight update should be negative. If the context is true and the node with it is selected, then a weight update should be positive.

## A2. Equations

There are also some equations that can measure how the concept trees should be used.

Tree shape can define a more stable wide and shallow tree, or a less stable narrow and deep one. This can be measured by comparing the total number of nodes to the tree depth or width.

$fs_t(\text{n, d})$. Equation 1

The decision to re-join trees may require intelligence that goes against the natural entropy decision to break a tree up. This decision can be based on the two shape functions, along with the branch link (count) in tree 1 ($l_{t1}$) and the base node link (count) in tree 2 ($l_{t2}$) and finally any concepts (context) in either tree ($C_{t1}$ and $C_{t2}$)

$f_{join}(fs_{t1}, fs_{t2}, l_{t1}, l_{t2}, C_{t1}, C_{t2})$. Equation 2

Adding new information to the concept base can be stored as a tuple of the concept value and a list of related descriptor values.

[A1, (Va1,Va2)] [B1, Vb1] [C1, (Vc1,Vc2,Vc3)] Equation 3